%% file: main.tex
\documentclass[10pt,twocolumn,letterpaper]{article}

\usepackage[left=0.8in,right=0.8in,top=0.8in,bottom=0.8in]{geometry}
\usepackage{cite}
\usepackage{amsmath,amssymb,amsfonts}
\usepackage{algorithmic}
\usepackage{graphicx}
\usepackage{textcomp}
\usepackage{xcolor}
\def\BibTeX{{\rm B\kern-.05em{\sc i\kern-.025em b}\kern-.08em
    T\kern-.1667em\lower.7ex\hbox{E}\kern-.125emX}}
\usepackage{booktabs}
\usepackage{blindtext}
\usepackage{hyperref}
\usepackage{comment}
\usepackage{subcaption}
\usepackage{xspace}
\usepackage{bm}
\usepackage{multirow}

\usepackage[capitalize]{cleveref}
\crefname{section}{Sec.}{Secs.}
\Crefname{section}{Section}{Sections}
\Crefname{table}{Table}{Tables}
\crefname{table}{Tab.}{Tabs.}

\makeatletter
\DeclareRobustCommand\onedot{\futurelet\@let@token\@onedot}
\def\@onedot{\ifx\@let@token.\else.\null\fi\xspace}

\def\eg{\emph{e.g}\onedot} 
\def\ie{\emph{i.e}\onedot}

\def\etal{\emph{et al}\onedot}
\makeatother

\definecolor{blue}{RGB}{0,50,200}
\definecolor{contrast-blue}{RGB}{52, 89, 237}
\definecolor{contrast-orange}{RGB}{230, 126, 0}
\definecolor{contrast-purple}{RGB}{121, 0, 201}

\begin{document}

\title{Almost Right: Making First-Layer Kernels \\Nearly Orthogonal Improves Model Generalization}

\author{
  Colton R. Crum \\
  University of Notre Dame \\
  \tt\small{ccrum@nd.edu}
  \and
  Adam Czajka \\
  University of Notre Dame \\
  \tt\small{aczajka@nd.edu}
}

\date{}

\maketitle

\input{latex/00-Abstract}
\input{latex/01-Introduction}
\input{latex/02-Related-Work}
\input{latex/03-Methodology}
\input{latex/04-Results}
\input{latex/05-Conclusion}

{\small
\bibliographystyle{ieee}
\bibliography{main}
}

\end{document}

%% file: latex/00-Abstract.tex
\begin{abstract}
Despite several algorithmic advances in the training of convolutional neural networks (CNNs) over the years, their generalization capabilities are still subpar across several pertinent domains, particularly within open-set tasks often found in biometric and medical contexts. On the contrary, humans have an uncanny ability to generalize to unknown visual stimuli. The efficient coding hypothesis posits that early visual structures (retina, Lateral Geniculate Nucleus, and primary visual cortex) transform inputs to reduce redundancy and maximize information efficiency. This mechanism of redundancy minimization in early vision was the inspiration for CNN regularization techniques that force convolutional kernels to be orthogonal. However, the existing works rely upon matrix projections, architectural modifications, or specific weight initializations, which frequently overtly constrain the network's learning process and excessively increase the computational load during loss function calculation. In this paper, we introduce a flexible and lightweight approach that regularizes a subset of first-layer convolutional filters by making them pairwise-orthogonal, which reduces the redundancy of the extracted features but at the same time prevents putting excessive constraints on the network. We evaluate the proposed method on three open-set visual tasks (anomaly detection in chest X-ray images, synthetic face detection, and iris presentation attack detection) and observe an increase in the generalization capabilities of models trained with the proposed regularizer compared to state-of-the-art kernel orthogonalization approaches. We offer source codes along with the paper.
\end{abstract}

%% file: latex/01-Introduction.tex
\section{Introduction}
\label{sec:Almost-Right-Introduction}

\subsection{Problem Definition}
An ongoing research area is how to effectively raise the generalization capabilities of neural networks, which is crucial from an AI safety perspective, particularly within domains such as biometrics, medical diagnostics, and anomaly detection. In each of these domains, models must appropriately respond to classes not previously seen during training, and efficiently extract meaningful features in scenarios where training data is limited and often heavily imbalanced across classes. Specifically, in the case of biometric presentation attack detection (PAD), it is infeasible to anticipate new presentation attack instruments. Similarly, diagnostic radiology requires significant generalization capabilities to identify abnormalities found within radiographic samples. For example, tumors vary in shape, size, and location depending on the pathology of a given patient. In summary, there are domains in which zero-shot approaches generalizing well to unknown anomalies are crucial for human health (or even life), or pertinent from the viewpoint of a country's security. Thus, the observed quest for neural network regularization methods not only allows models to learn more efficiently but also to generalize better to unknown inputs.

\subsection{Proposed Solution}

Given that humans, animals, and even children have an exceptional ability to generalize to unknown stimuli and environments, research within model generalization is often inspired by findings from biology, cognitive science, and neuroscience communities \cite{boyd2021cyborg, crum2023mentor, linsley2018learning}. One of such inspirations is the efficient coding hypothesis for early vision (mainly achieved by appropriate construction of receptive fields in retina, lateral geniculate nucleus, and primary visual cortex) \cite{barlow1961possible,laughlin1981simple,atick1990early,dong1995statistics} and the ability to process and decompose signals from raw stimulus efficiently for subsequent downstream processing \cite{Koch2004}. Motivated by these theoretical neuroscience studies, there is a series of works in computer science exploring orthogonality of filtering kernels present in convolutional neural networks (CNNs), or de-correlation of information generated by these kernels \cite{wang2020orthogonal, xie2017all, bansal2018can, xu2022lot}. 

However, the existing works put various regularization constraints requesting orthogonality between the {\it full set} of convolutional kernels. This, as demonstrated in this paper, may not be an optimal solution since the number of kernels in contemporary CNNs is often arbitrary. Instead, in this paper, we propose a loss function-based and architecture-agnostic \emph{Almost Right} regularization, which requests the {\it selected pairs} of convolutional kernels within the first convolutional layer to be nearly orthogonal. Contrary to previous works, the learning process has the flexibility to choose (i) the kernels that are orthogonalized, and (ii) the strength of such orthogonalization, leading to significant performance gains. Although seemingly nuanced, this pairwise orthogonalization proves crucial in allowing the network to focus on the main body of kernels that can be made orthogonal, which -- surprisingly -- significantly increases the network's generalization capabilities.

\subsection{Research Questions and Contributions}
The paper is organized around four research questions:

\begin{itemize}
    \item \textbf{RQ1}: Can the {\it Almost Right} regularization, encouraging the model to make {\bf pairs} of first-layer convolutional kernels orthogonal, raise model generalization?
    
    \item \textbf{RQ2}: How does {\it Almost Right} compare against state-of-the-art kernel orthogonalization regularizers?

    \item \textbf{RQ3}: How does {\it Almost Right} compare against existing state-of-the-art saliency-based regularizers?

    \item \textbf{RQ4}: Can {\it Almost Right} generalize across vision architectures, including both CNNs and vision transformers?

\end{itemize}

We evaluate the proposed \emph{Almost Right} loss within both open-set and closed-set scenarios across six unique datasets and five architectures, including CNNs and Vision Transformers (ViT). We also offer source codes of the proposed method to facilitate future research in this area.

%% file: latex/02-Related-Work.tex
\section{Related Work}
\label{sec:Almost-Right-Related-Work}

\paragraph{Biological Inspirations:}

Regularization of model training utilizing kernel orthogonalizations and decorrelation of processed information (e.g., in feature maps) takes inspiration from neuroscience and the understanding of early human vision. In particular, the first module of the visual cortex (V1) plays a pivotal role in how humans see and perceive the outside world \cite{Koch2004, deangelis2004modern, deangelis1995receptive}, and is the brain's first attempt at filtering, decomposing, and efficiently encoding visual information captured by the eyes for subsequent neurons downstream \cite{barlow1961possible,laughlin1981simple,atick1990early,dong1995statistics}.

Studies proposing a two-dimensional Gabor filter as a good model of simple receptive fields in the first cortical areas \cite{daugman1980two,Jones_Neurophysiology_1987} inspired works, in which Gabor wavelets were proposed to serve as effective feature extractors, including biometric recognition \cite{Daugman_TASSP_1988, daugman1985uncertainty,Daugman_PAMI_1993} or adversarial robustness fields \cite{dapello2020simulating}. Although Gabor wavelets are not orthogonal, they possess an interesting property of being capable of disentangling spatial frequency and orientation and serve as an inspiration for a larger set of regularization techniques that aim at efficient representations of visual information.

\paragraph{Kernel Orthogonalization:} Existing methods proposed to date focus on regularizing the network by imposing explicit or implicit orthogonality on the entire set of kernels of the network \cite{xie2017all, wang2020orthogonal, saxe2013exact, mishkin2015all}. These methods utilize orthogonality to decorrelate filtering results within a convolutional layer, which can reduce information redundancy within the network \cite{mishkin2015all}, thereby increasing its classification performance \cite{wang2020orthogonal} and reducing instabilities during training \cite{saxe2013exact, xie2017all}.
Additionally, Saxe \etal demonstrated that initializing networks with orthonormal weights can lead to faster convergence compared to random Gaussian weights \cite{saxe2013exact}. Building upon their work, Mishkin \etal introduced layer-sequential unit-variance (LSUV), an iterative initialization procedure where each convolution or inner-product layer is initialized with orthonormal matrices, followed by a normalization of the variance of every layer in the network \cite{mishkin2015all}. Other works have attempted to integrate orthogonality through architectural modifications \cite{wang2020orthogonal}, initialization methods \cite{xie2017all, saxe2013exact, mishkin2015all}, or enforcing orthogonality upon convolutional kernels \cite{bansal2018can}. Wang \etal proposed Orthogonal Convolutional Neural Networks (OCNN), which imposed orthogonal convolutions based upon doubly block-Toeplitz matrix representations \cite{wang2020orthogonal}.

\paragraph{Research Gaps:}
Nearly all previous works require stringent architectural and training configurations that restrict practical implementation \cite{wang2020orthogonal, saxe2013exact, xie2017all}. Within initialization techniques, training configurations play a crucial role in network convergence, which is often predicated upon the target domain and the availability of rich training samples \cite{saxe2013exact, xie2017all, mishkin2015all}. Thus, orthogonal initialization techniques have proven difficult within open-set scenarios where training data is often class-imbalanced, underrepresented, and not well understood in advance.

Network architectures also play a significant role in calculating orthogonality due to nuances related to layer dimensionality \cite{bansal2018can, wang2020orthogonal}. As a consequence, existing orthogonality methods often rely on modification or projection of kernel sizes and channels to satisfy these constraints \cite{bansal2018can}. Orthogonal neural networks, which are intended to smooth frequencies and regularize model training often, induce a host of hyper-parameters to govern their successful use, which is a rather counterintuitive byproduct of these works \cite{wang2020orthogonal}.

\paragraph{How this paper differs from previous studies?} Rather than relying on strict orthogonality constraints for all network kernels, we propose a new loss component that can be added to the training of any CNN, which encourages the network to maximize the angle between selected pairs of kernels. This flexible approach naturally adapts to different kernel numbers and sizes, and allows the model to benefit from an efficient encoding paradigm without over-regulating its training. The result, as this paper demonstrates, is better generalization across different architectures and tasks.

%% file: latex/03-Methodology.tex
\section{Soft Kernel Orthogonalization}
\label{sec:Almost-Right-Methodology}

This section introduces the proposed ``nearly orthogonal'' loss component. First, we delineate our intuition beyond that specific loss construction and demarcate several novel differences with respect to previous works. Second, based upon those considerations, we formally introduce a softly orthogonalizing loss. Finally, we describe the procedure to evaluate the performance of the proposed method.

\subsection{Summary of the Approach}
Our method is inspired by elements of previous works, imposing orthogonal regularization directly upon network training \cite{bansal2018can, wang2020orthogonal}. However, we novelly deviate from prior works by: (a) considering all channels of the first layer only, (b) considering only pairs of kernels instead of all kernels simultaneously, and (c) allowing some flexibility to encourage \emph{near} orthogonality between vector pairs, enabling some unique pairings to have angles different than $90^\circ$. We elucidate each of these points in the following paragraphs.

\paragraph{Flattening of All Channels:} The applicability of the proposed loss function to any CNN architecture is driven by the flattening of color channels across all dimensions. Our intuition is that each color channel is intermixed with one another in uninterpretable ways after the first layer of the network. To the best of our knowledge, all pretrained visual classification architectures available in PyTorch have a 2-dimensional convolutional layer as the first layer of the network \cite{paszke2017automatic}. As a consequence, we flatten each convolutional kernel along each color channel and dimension, which allows significant implementation flexibility.

Moreover, the number of output channels (= kernels) of the first layer is usually arbitrary and is sometimes a relic of an experimental design procedure of the past. Although the number of channels is indeed important, there is no single ``goldilocks'' value for output channels unless extensive network architecture search or hyper-parameter optimization on a dataset and task-specific basis.

\paragraph{Pairwise Neuron Comparisons:} We consider pairwise orthogonalization of kernels, instead of forcing all filters (channels) to be orthogonal. This, in addition to letting the model ``choose'' during training which kernels should be orthogonal, also allows for flexibility in implementation, as considering all kernels requires the number of kernels equal to the flattened size of the kernels (to operate on square matrices in orthogonalization loss components). Consequently, existing methods often require an architectural modification or a linear projection step to reconfigure the dimensionality of outputs in order to put an orthogonality constraint on the entire set of kernels \cite{wang2020orthogonal, bansal2018can}.

\paragraph{Soft Orthogonality:} We hypothesize that encouraging \emph{near} orthogonal $N$-dimensional vectors as opposed to strictly orthogonal will also increase the learning capacity of the network, purported by the Johnson-Lindenstrauss lemma \cite{johnson1984extensions}. The lemma states that when vectors are \emph{nearly} orthogonal with one another, the number of vectors that can fit within the $N$-dimensional subspace grows exponentially, allowing the network to compress more information within a given manifold simply by slight adjustments to their angles \cite{larsen2017optimality, alon2003problems, levenshtein1983bounds}.
\subsection{Loss Formulation}

Taking the above factors into consideration, we introduce our {\it Almost Right} loss component as:

\begin{equation}
\begin{split}
\mathcal{L_\text{Almost Right}} = \frac{2}{K(K-1)}\sum_{i=1}^K\sum_{j>i}^K {\frac{\vec{A_i}\cdot\vec{A_j}}{{(||\vec{A}_i}||\cdot||\vec{A_j}|| +\epsilon)}}
\end{split}
\end{equation}

\noindent where $A_i$ is the $i$-th kernel, $K$ is the total number of kernels in a given layer, $||\vec{A}||$ is the length of the flattened 2D kernel, that is: (number of input channels)$\times$(kernel size)$^2$, and $\epsilon=1e$--$08$ is added to avoid division by zero.

We then combine a regular cross-entropy loss:
\begin{equation}
\begin{split}
\mathcal{L}_\text{cross entropy} = \sum_{c=1}^{C}\bm{1}_{y \in \mathcal{C}_c} \log p\big(y \in \mathcal{C}_c\big)
\end{split}
\end{equation}

\noindent where $y$ is a class label, $\bm{1}$ is a class indicator function equal to $1$ when $y \in \mathcal{C}_c$ (and 0 otherwise), $C$ is the total number of classes, with the proposed soft kernel orthogonalization term:

\begin{equation}
\label{eq:almost-ar-loss}
\mathcal{L} = \alpha\mathcal{L}_\text{cross entropy} + (1-\alpha)\mathcal{L}_\text{Almost Right}
\end{equation}

\noindent where $\alpha$ is a hyperparameter weighting the nearly orthogonal and cross-entropy loss terms. Unless explicitly stated, we perform no hyper-parameter optimization and select equal weighting of both loss components (\ie, $\alpha=0.50$).

\section{Methodology}

\subsection{Datasets}

We evaluate the proposed method against the existing orthogonal regularization approaches and architectures \cite{bansal2018can, wang2020orthogonal}, and across three open-set domains: iris presentation attack detection (PAD), anomaly detection in chest X-ray radiographs, and synthetic face detection. The three selected state-of-the-art datasets offer human annotations, which allows us to also compare the proposed regularization with selected saliency-based training methods \cite{boyd2021cyborg, crum2023mentor, karargyris2021creation}, whose goal is the same: increase the model's generalization within an open-set recognition scenario.

\input{figures/dataset-sampler-open-set}

\paragraph{Domain 1 -- Iris Presentation Attack Detection:} Training and validation samples consisting of {\it{bona fide}} and spoof iris samples were selected from a superset of previously published live iris and iris presentation attack samples \cite{casia-database,Sung_OE_2007,Galbally_ICB_2012,Kohli_ICB_2013,Yambay_ISBA_2017,Trokielewicz_IVC_2020,Kohli_BTAS_2016,Wei_ICPR_2008,Trokielewicz_BTAS_2015,Yambay_IJCB_2017,das2020iris}. Presentation attack instruments represented in the dataset include postmortem, artificial (glass or doll eye), textured contact lens, paper printouts, textured contact lens printouts, and synthetically generated iris samples. We use the Iris Liveness Detection 2020 Competition \cite{das2020iris} as our test set (disjoint with training and validation sets) following previous works \cite{crum2023mentor, crum2024grains}.

\paragraph{Domain 2 -- Chest X-ray Scanning:} Training, validation, and test disjoint sets were sampled from chest X-rays provided by Johnson \etal \cite{johnson2019mimic}. For consistency, we maintain the same number of training samples as the iris PAD domain. Abnormal samples include at least one abnormality, consisting of pneumonia (infection), atelectasis (complete or partial lung collapse), pleural effusion (fluid buildup), cardiomegaly (enlarged heart), edema (swelling), lung opacity (abnormal tissue density), or support devices (\eg pacemakers, etc.). Normal samples have no visible abnormalities as confirmed by several board-certified radiologists. The test set consists of 8,265 images.

\paragraph{Domain 3 -- Synthetic Face Detection:} The training set consists of 919 real and 902 synthetically-generated face images selected from previously published works \cite{karras2017progressive, stargan, karras2020analyzing, Karras2020ada, karras2021alias}, following Boyd \etal \cite{boyd2021cyborg}. Real faces were sampled from FFHQ \cite{karras2017progressive} and Celeb-HQ \cite{karras2017progressive}, whereas synthetically generated faces were generated from various StyleGAN generators. The validation set, disjoint from the training set, comprises of 20,000 images extracted from the same sources. The test set consists of $7,000$ images: $2,000$ real face samples evenly sourced from Celeb-HQ \cite{karras2017progressive} and FFHQ \cite{karras2017progressive}, and $5,000$ evenly sourced synthetic face samples from ProGAN \cite{karras2017progressive}, StyleGAN \cite{karras2017progressive}, StyleGAN2 \cite{karras2020analyzing}, StyleGAN2-ADA \cite{Karras2020ada}, StyleGAN3 \cite{karras2021alias}.

For all domains featured in this paper, all training, validation, and testing sets are disjoint and follow the same evaluation procedures and splits as in prior works \cite{boyd2021cyborg,crum2023mentor} to offer direct comparisons with the state-of-the-art. Each task is considered open-set since not all potential spoofing attack types (iris PAD domain), abnormalities (chest X-rays domain), and fake sample generators (synthetic face detection domain) are seen during training and validation. These domains also reflect heavy class imbalances (\eg 198 live / 567 spoof samples for iris PAD, 223 normal / 542 abnormal samples for chest X-ray anomalies, 919 authentic / 902 synthetic for synthetic face detection) and contain a limited number of training samples. Such imbalances and low sample availability are natural in biometric safety and healthcare domains, so this introduces real-world complications to our evaluations.

\subsection{Baseline Methods}

\paragraph{Orthogonal Regularizers:} We compare the proposed method against several state of the art orthogonal regularization methods. Architectural-based methods include Orthogonal Convolutional Neural Networks (OCNN) \cite{wang2020orthogonal} and Orthogonal Regularization introduced by Bansal \etal \cite{bansal2018can}. We also include an orthogonal-based weight initialization approach, Layer-Sequential Unit-Variance Initialization (LSUV Init) \cite{mishkin2015all}, which is a generalization of earlier work \cite{saxe2013exact}.

\paragraph{Saliency-based Regularizers:} We demonstrate that the proposed method can perform competitively with, or better than saliency-based regularization methods, which rely upon the use of human annotations to guide the model during training toward salient features. We include methods considered state of the art for the open-set problems: CYBORG \cite{boyd2021cyborg}, Unet+Gaze \cite{karargyris2021creation}, and MENTOR \cite{crum2023mentor}. In these previous works, human subjects were asked to annotate salient regions of the input image using mouse annotations for iris PAD and synthetic face detection domains \cite{boyd2021cyborg, boyd2022human}. For chest X-ray abnormalities, saliency maps were created from eye-tracking data provided by several board-certified radiologists \cite{johnson2019mimic, karargyris2021creation}.

\paragraph{Wavelet-based Regularizer:} We also include VOneNet, a Gabor wavelet-based regularizer, due to its emphasis on regularizing the first layer of the network \cite{dapello2020simulating}. Though this regularization focused on increasing the adversarial robustness of CNNs, rather than open-set recognition, it is based on a neuroscientific model of the primary visual cortex, and it incorporates Gabor wavelets. This family of wavelets possesses an important property of being optimal within the uncertainty principle, that is, maximizing the simultaneous space–frequency resolution and thus offering the best precision in extracting visual features as defined by Aristotle's concept of vision (``knowing what is where'').

All methods are implemented using the source codes released by the authors.

\subsection{Training Configurations \& Evaluation Metrics}
\label{sec:almost-training-config}
We follow previous works employing the same datasets and use the same training configurations to facilitate direct comparisons with results from the existing literature \cite{boyd2021cyborg, crum2023mentor, crum2024grains, crum2023teaching}. All models are initialized with ImageNet weights, trained using a batch size of 20 for 50 epochs with SGD, with an initial learning rate of 0.005 and a scheduler that reduces the learning rate by 0.1 every 12 epochs. The weighted sum between cross-entropy and the nearly orthogonal loss component remained fixed at $\alpha=0.5$.

We evaluate our method in open-set scenarios using Area Under the Receiver Operating Characteristic Curve (AUROC). We carry out multiple independent train-test runs to offer the statistical assessment of uncertainty (as a standard deviation from the mean) of the observed results.

\subsection{Network Architectures}
Prior works in orthogonal regularization focus almost exclusively on residual-based networks due to stringent architecture requirements \cite{wang2020orthogonal, bansal2018can}. Conversely, we highlight the flexibility of the proposed method by including extensive experiments across several architectures and model sizes (\ie number of trainable parameters), including both CNNs and vision transformers: ResNet50 \cite{he2016deep}, DenseNet-121 \cite{huang2017densely}, ConvNext-Tiny \cite{liu2022convnet}, Vit base-16 \cite{dosovitskiy_image_2021}, and Swin-Tiny \cite{liu2021swin}.

Our method is applicable to \emph{any} neural network (including vision transformers) that uses a convolution operation in the first layer. Interestingly, we found that -- when the proposed orthogonalization regularization is active -- replacing the ReLU with the Sigmoid activation function after the first convolutional layer helps in a better flow of information to the remainder of the layers. Our findings are consistent with previous works in orthonormal initialization, which noted how batch normalization preceding a ReLU activation function effectively removes half of the activations from the network \cite{xie2017all}. We leave various experimental considerations like this one for interesting future work.

%% file: figures/dataset-sampler-open-set.tex
\begin{figure}[t!]
  \centering
  \includegraphics[width=\linewidth]{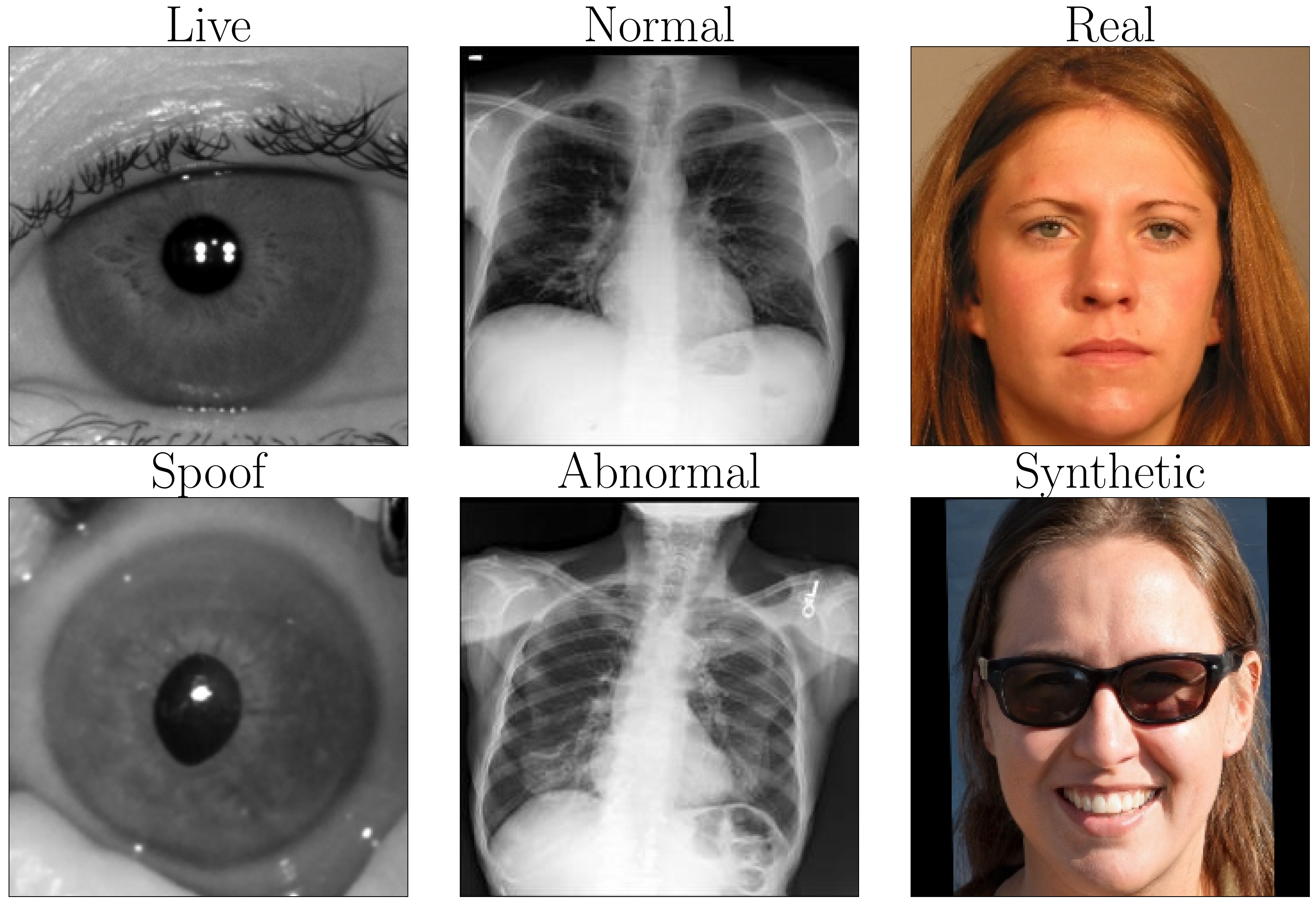}
  \caption{A sample of the three open-set recognition tasks evaluated in this paper: \textbf{Iris presentation attack detection (first column)}, \textbf{Chest X-ray abnormalities (middle)}, and \textbf{Synthetic face detection (third)}. The first row indicates \textbf{live} (Iris PAD), \textbf{normal} (Chest X-rays), and \textbf{real} samples, whereas the bottom row indicates \textbf{spoof}, \textbf{abnormal}, and \textbf{fake} samples.}
  \label{fig:almost-dataset}
\end{figure}

%% file: latex/04-Results.tex
\section{Results}
\label{sec:Almost-Right-Results}

This section is organized to clearly answer the research questions listed in the Introduction. First, we demonstrate the effectiveness of our method in raising the generalization capabilities compared to the baseline. Second, we compare these gains to SOTA orthogonal-based regularization methods. Next, we explore how our method performs against SOTA saliency-based regularizers. Finally, we illustrate the flexibility of our method by evaluating it with the most popular convolutional neural network architectures.

\input{tables/open-set}

\subsection{Addressing RQ1 on Open-set Generalization}
Table \ref{tab:almost-open-set} presents aggregated results (means and standard deviations of AUROC scores calculated across 5 independent train-test runs) for all three datasets and various regularizers built for several model architectures. As it can be seen, {\bf the proposed {\it Almost Right} approach raises the generalization performance in all three open-set recognition tasks}. 
For the ResNet architecture, used across all tested methods, {\it Almost Right} increased its baseline generalization performance from 0.878 to 0.913, 0.832 to 0.848, and 0.474 to 0.572 for iris PAD, chest X-ray abnormalities detection, and synthetic face detection, respectively.

\subsection{Addressing RQ2 on Open-set Performance Against SOTA Orthogonal Regularizers}
The proposed {\it Almost Right} posts strong generalization compared to various orthogonal-based regularizers. For iris PAD, it outperforms existing orthogonal-based regularizers by a wide margin (0.913). Following {\it Almost Right} is OCNN (0.880), with VOneNet (0.686), Ortho. Reg (0.678), and LSUV (0.605) trailing significantly behind, even compared to the Baseline (CE) where no regularization is applied. For the Chest X-ray anomaly detection task, {\it Almost Right} holds strong performance (0.848), followed closely by OCNN (0.846). VOneNet (0.786), LSUV (0.774), and Ortho. Reg. (0.654) all struggled to outperform the baseline. Finally, our method yields some of the most substantial gains in synthetic face detection. {\it Almost Right} leads by a wide margin (0.572), where the other methods failed to surpass random chance: OCNN (0.482), Ortho. Reg. (0.425), VOneNet (0.474), and LSUV (0.427). We hypothesize that the performance gain of {\it Almost Right} is due in part to the regularization of the high-frequency signatures contained within synthetic samples, which was noted as a challenge in previous works \cite{crum2023explain}.

These results may also suggest that the synthetic face detection task has more correlated features in its training data compared to the iris PAD and chest X-ray anomalies detection datasets. Synthetic face samples were all generated by StyleGAN-based architectures trained on the same data, further reducing their sample diversity. On the contrary, iris PAD and chest X-ray samples offer a more diversified training set, where abnormal samples were highly diverse across at least six different sources. Thus, we observe the largest performance gains in synthetic face detection since our method encourages the decorrelation of first-layer kernels, which facilitates more efficient feature extraction in cases where the representation of the training set is sparse.

Finally, it is worth mentioning that previous works in orthogonal regularization were not evaluated in open-set scenarios, which helps explain the reduction in performance as reported in the original papers \cite{wang2020orthogonal, dapello2020simulating,mishkin2015all, bansal2018can}. Moreover, OCNN was evaluated using validation accuracy \cite{wang2020orthogonal} instead of separate test sets, which is unrealistic and not applicable in the scope of our experiments.

\subsection{Addressing RQ3 on Open-set Performance Against SOTA Saliency-based Regularizers}
Unlike saliency-based training methods, {\it Almost Right} uses no supplemental human-perceptual information during training and still surpasses their performance with often comfortable margins. CYBORG and Unet+Gaze are loss-based saliency-based regularizers \cite{boyd2021cyborg, karargyris2021creation}, whereas MENTOR implements a two-stage training strategy \cite{crum2023mentor}. For the iris PAD task, {\it Almost Right} achieves an average AUROC=0.913, with the nearest competitor being MENTOR (0.909), with CYBORG (0.886), and Unet+Gaze (0.861) following further behind.

Within synthetic face detection, {\it Almost Right} surpasses SOTA saliency-based methods by a fair margin (0.572), followed by CYBORG (0.531). MENTOR (0.528), baseline cross-entropy (0.474), and Unet+Gaze (0.490) all perform at random chance on average.

\subsection{Addressing RQ4 on Performance Across Architectures}

The proposed {\it Almost Right} method can be applied across different architectures, including both CNNs and Vision Transformers. Unlike many orthogonalization methods, which often require specific architectures or layer-wise modifications due to dimensionality requirements,
our method can be implemented across any architecture that implements a convolution in the first layer. Technically, this encompasses all classification models currently available on PyTorch.

As can be seen from Tab. \ref{tab:almost-open-set}, {\it Almost Right} achieved the top performance for nearly all model architectures and experimental configurations. Specifically, the performance increase for ConvNext in iris and chest X-ray anomalies detection was significant (61\% and 35\%, respectively). Interestingly, we observe the highest gains over the baseline for vision transformer architectures: {\it Almost Right} boosted the generalization performance for Vit-b by 13\% (iris PAD task), 14\% (chest X-ray anomaly detection), and 52\% (synthetic face detection). These results are also consistent with the Swin-T transformer architecture. These results may suggest that vision transformers may have more redundant convolutional filters within the first layer of the network. Methods such as the proposed {\it Almost Right} can effectively regularize such architectures, allowing them to improve performance in open-set classification scenarios.

We hypothesize that large performance gains observed for ConvNext, Vit-b, and Swin-T architectures may also be due in part to the location of the batch normalization coupled with ReLU activation functions. The rescaling of filters followed by the removal of negative values from ReLU likely works against the orthogonalization of the first-layer filters. Unlike ResNet and DenseNet, which contain a batch normalization layer immediately following the first convolutional layer, ConvNext, Vit-b, and Swin-T instead have at least one layer (or more) in between the convolution and the batch normalization layer. These observations are consistent with previous work in orthogonal weight initialization methods \cite{xie2017all}.

Finally, a small number of architectures achieved lower performance gains for different reasons. For the chest X-ray anomaly detection task, {\it Almost Right} (0.848) holds marginal performance gains compared to existing methods across DenseNet and Vit-b backbones. We hypothesize that this may be due in part to the DenseNet's and Swin transformer's distinctive use of cross-network connections (\ie skip connections in dense blocks \cite{huang2017densely} and cross-window connections \cite{liu2021swin}), allowing information to move more freely across the network and thereby reducing the effect of first-layer regularization.

\section{Ablation Study}

We also include an ablation study to explore (a) how the {\it Almost Right} loss performs within closed-set classification scenarios, where all evaluation classes are present during training, and (b) the sensitivity of the generalization performance gain to regularization strength.

\input{tables/closed-set}

\subsection{Closed-set Classification Performance}
We explore our method across three general vision datasets in a closed-set classification scenario: Oxford Flowers \cite{4756141}, Food \cite{wu2021large}, and Caltech-CUB-2011 \cite{wah2011caltech}. Unlike the previous datasets used in an open-set classification scenario, these datasets are class-balanced, designed for multi-class classification (as opposed to binary classification), and indicative of more generic vision applications of our method. Where applicable, validation sets are created using an 80/20 split from the training data. For consistency with prior experiments, we use the same training configurations as the previous experiments (see Sec. \ref{sec:almost-training-config}). Test sets are disjoint from training and validation sets. 

We evaluate each method using top-1 mean accuracy, with standard deviations reported across 5 independent runs. As shown in Tab. \ref{tab:almost-closed-set}, our method has a fair to marginal improvement in closed-set classification performance depending on the architecture and dataset. For the Food dataset, {\it Almost Right} has substantial performance gains across ResNet backbones (0.703) compared to the Baseline (0.666) and OCNN (0.586). LSUV and Ortho. Reg. struggled to converge even after repeated training runs, and as a consequence had poor performance (0.161 and 0.144, respectively). These findings are consistent across nearly all architectures and methods explored in this paper. Notably, our method boosted performance with respect to the Baseline for ConvNext (0.721 $\rightarrow$ 0.757) and Swin-T (0.747 $\rightarrow$ 0.779). {\it Almost Right} was only second in one experiment with Vit-b, and was matched by CYBORG for ResNet. For the Flowers and Caltech CUB 2011 datasets, {\it Almost Right} has marginal or mixed improvement depending on the architecture. Our method matched the performance of the Baseline for ResNet architectures in both Flowers and CUB. For DenseNet, we performed modestly for Flowers (0.885) and CUB (0.752), scoring behind the Baseline but frequently ahead of other regularization methods. These results are consistent for Vit-b across both Flowers and CUB. Similar to Food, our method posted the top performance in the other datasets for ConvNext and Swin architectures.

Concluding this study, our method shows a fair performance improvement compared to the findings of previous orthogonal-based works \cite{wang2020orthogonal, bansal2018can, xie2017all, mishkin2015all}. However, greater gains can be made within open-set classification scenarios, which is the primary focus of this paper. We hypothesize that soft orthogonalization is more advantageous within open-set due in part to imbalances within the anomalous sample subset. Given that our loss leaves the network unbounded in terms of how far it orthogonalizes the first layer, our results suggest there may be a well-defined ``goldilocks'' zone for near orthogonalization. Lower performance gains in the closed-set classification scenario may also be a consequence of Food, Flowers, and CUB datasets being well-represented and more class-balanced.

\subsection{Exploring Regularization Strength}

\input{tables/alpha-open-set}
In this ablation study, we explore how changing the regularization strength within the proposed loss (controlled by $\alpha$ in Eq. \eqref{eq:almost-ar-loss}) may impact the generalization performance (see Tab. \ref{tab:almost-ablation-alpha}). Alpha controls the weighting between orthogonal and cross-entropy loss components, where lower values indicate more orthogonal regularization and higher values indicate less. The results from Tab. \ref{tab:almost-ablation-alpha}, presented for ResNet backbone and open-set classification tasks, suggest that for nearly all tasks the choice of $\alpha=0.5$ offers a good balance between the two loss components.

%% file: tables/open-set.tex
\begin{table*}[t!]
\footnotesize
\centering
\caption{Evaluation across {\bf open-set classification scenarios}: Iris (Iris Presentation Attack Detection), Chest (Chest X-ray Anomaly Detection), and Face (Synthetic Face Detection). Means and standard deviations of AUROC scores are reported across 5 independent runs. The best method for each dataset and architecture is \textbf{bolded}.}
\vspace{-3mm}
\label{tab:almost-open-set}
\begin{tabular}{l|l|ccc}
\toprule
\textbf{Method} & \textbf{Backbone}& \textbf{Iris Presentation} & \textbf{Chest X-ray} & \textbf{Synthetic Face} \\
& & \textbf{Attack Detection} & \textbf{Anomaly Detection} & \textbf{Detection} \\
\midrule
Baseline & ResNet & 0.878\tiny{$\pm$0.018} & 0.832\tiny{$\pm$0.009} & 0.474\tiny{$\pm$0.080} \\
(cross entropy only) & DenseNet & 0.881\tiny{$\pm$0.016} & 0.829\tiny{$\pm$0.015} & 0.508\tiny{$\pm$0.042} \\
 & ConvNext & 0.528\tiny{$\pm$0.044} & 0.629\tiny{$\pm$0.089} & 0.428\tiny{$\pm$0.032} \\
 & Vit-b & 0.609\tiny{$\pm$0.027} & 0.684\tiny{$\pm$0.061} & 0.450\tiny{$\pm$0.029} \\
 & Swin-T & 0.657\tiny{$\pm$0.071} & 0.777\tiny{$\pm$0.059} & 0.472\tiny{$\pm$0.018} \\
\midrule
LSUV \cite{mishkin2015all} & ResNet & 0.605\tiny{$\pm$0.020} & 0.774\tiny{$\pm$0.017} & 0.427\tiny{$\pm$0.044} \\
 & DenseNet & 0.890\tiny{$\pm$0.014} & 0.835\tiny{$\pm$0.017} & 0.507\tiny{$\pm$0.057} \\
 & ConvNext & 0.521\tiny{$\pm$0.015} & 0.783\tiny{$\pm$0.090} & 0.429\tiny{$\pm$0.036} \\
 & Vit-b & 0.617\tiny{$\pm$0.030} & 0.727\tiny{$\pm$0.039} & 0.429\tiny{$\pm$0.020} \\
 & Swin-T & 0.619\tiny{$\pm$0.074} & 0.791\tiny{$\pm$0.041} & 0.486\tiny{$\pm$0.040} \\
 \midrule
 OCNN \cite{wang2020orthogonal} & ResNet & 0.880\tiny{$\pm$0.008} & 0.846\tiny{$\pm$0.007} & 0.482\tiny{$\pm$0.051} \\
\midrule
Orthogonality Regularization \cite{bansal2018can} & ResNet & 0.678\tiny{$\pm$0.046} & 0.654\tiny{$\pm$0.137} & 0.425\tiny{$\pm$0.040} \\
\midrule
VOneNet \cite{dapello2020simulating} & ResNet & 0.686\tiny{$\pm$0.025} & 0.786\tiny{$\pm$0.026} & 0.474\tiny{$\pm$0.080} \\
\midrule
CYBORG \cite{boyd2021cyborg} \ddag & ResNet & 0.886\tiny{$\pm$0.020} & 0.840\tiny{$\pm$0.019} & 0.531\tiny{$\pm$0.050} \\
\midrule
Unet+Gaze \cite{karargyris2021creation} \ddag & ResNet & 0.861\tiny{$\pm$0.019} & 0.835\tiny{$\pm$0.013} & 0.490\tiny{$\pm$0.054} \\
 & DenseNet & 0.860\tiny{$\pm$0.013} & 0.828\tiny{$\pm$0.006} & 0.505\tiny{$\pm$0.030} \\
 & ConvNext & 0.510\tiny{$\pm$0.015} & 0.631\tiny{$\pm$0.043} & \textbf{0.441\tiny{$\pm$0.016}} \\
 & Swin-T & 0.484\tiny{$\pm$0.018} & 0.681\tiny{$\pm$0.069} & 0.463\tiny{$\pm$0.007} \\
\midrule
MENTOR \cite{crum2023mentor} \ddag & ResNet & 0.909\tiny{$\pm$0.023} & 0.844\tiny{$\pm$0.006} & 0.528\tiny{$\pm$0.061} \\
 & DenseNet & 0.893\tiny{$\pm$0.008} & \textbf{0.841\tiny{$\pm$0.009}} & 0.565\tiny{$\pm$0.035} \\
 & ConvNext & 0.722\tiny{$\pm$0.044} & 0.824\tiny{$\pm$0.009} & 0.425\tiny{$\pm$0.022} \\
 & Swin-T & 0.698\tiny{$\pm$0.079} & 0.834\tiny{$\pm$0.013} & 0.439\tiny{$\pm$0.014} \\
\midrule
{\bf Almost Right} & ResNet & \textbf{0.913\tiny{$\pm$0.012}} & \textbf{0.848\tiny{$\pm$0.006}} & \textbf{0.572\tiny{$\pm$0.032}} \\
{\bf (proposed)} & DenseNet & \textbf{0.891\tiny{$\pm$0.016}} & \text{0.836\tiny{$\pm$0.009}} & \textbf{0.560\tiny{$\pm$0.050}} \\
 & ConvNext & \textbf{0.850\tiny{$\pm$0.025}} & \textbf{0.848\tiny{$\pm$0.008}} & 0.427\tiny{$\pm$0.043} \\
 & Vit-b & \textbf{0.685\tiny{$\pm$0.032}} & \textbf{0.778\tiny{$\pm$0.028}} & \textbf{0.685\tiny{$\pm$0.032}} \\
 & Swin-T & \textbf{0.729\tiny{$\pm$0.026}} & \textbf{0.834\tiny{$\pm$0.010}} & \textbf{0.506\tiny{$\pm$0.020}} \\
\bottomrule
\end{tabular}
\end{table*}

%% file: tables/closed-set.tex
\begin{table*}[t!]
\centering
\footnotesize
\caption{Same as Tab. \ref{tab:almost-open-set}, except methods are evaluated in {\bf closed-set classification scenarios} and top-1 accuracy is reported.}
\vspace{-3mm}
\label{tab:almost-closed-set}
\begin{tabular}{l|l|ccc}
\toprule
\textbf{Method} & \textbf{Backbone} & \textbf{Food} & \textbf{Oxford Flowers} & \textbf{Caltech CUB 2011} \\

\midrule
Baseline & ResNet & 0.666\tiny{$\pm$0.006} & \textbf{0.891\tiny{$\pm$0.002}} & \textbf{0.744\tiny{$\pm$0.005}} \\
(cross entropy only) & DenseNet & 0.707\tiny{$\pm$0.006} & \textbf{0.890\tiny{$\pm$0.005}} & \textbf{0.768\tiny{$\pm$0.004}} \\
 & ConvNext & 0.721\tiny{$\pm$0.025} & 0.727\tiny{$\pm$0.095} & 0.763\tiny{$\pm$0.007} \\
 & Vit-b & 0.609\tiny{$\pm$0.003} & 0.748\tiny{$\pm$0.047} & 0.630\tiny{$\pm$0.007} \\
 & Swin-T & 0.747\tiny{$\pm$0.009} & 0.952\tiny{$\pm$0.013} & 0.788\tiny{$\pm$0.006} \\
\midrule
LSUV \cite{mishkin2015all} & ResNet & 0.161\tiny{$\pm$0.010} & 0.276\tiny{$\pm$0.008} & 0.142\tiny{$\pm$0.020} \\
 & DenseNet & 0.707\tiny{$\pm$0.008} & 0.888\tiny{$\pm$0.006} & \textbf{0.770\tiny{$\pm$0.003}} \\
 & ConvNext & 0.726\tiny{$\pm$0.010} & 0.425\tiny{$\pm$0.383} & 0.761\tiny{$\pm$0.004} \\
 & Vit-b & 0.579\tiny{$\pm$0.033} & 0.619\tiny{$\pm$0.099} & 0.515\tiny{$\pm$0.247} \\
 & Swin-T & 0.742\tiny{$\pm$0.005} & 0.944\tiny{$\pm$0.007} & 0.789\tiny{$\pm$0.007} \\
 \midrule
OCNN \cite{wang2020orthogonal} & ResNet & 0.586\tiny{$\pm$0.014} & 0.890\tiny{$\pm$0.002} & 0.731\tiny{$\pm$0.002} \\
\midrule
Orthogonality Regularization \cite{bansal2018can} & ResNet & 0.144\tiny{$\pm$0.008} & 0.348\tiny{$\pm$0.010} & 0.167\tiny{$\pm$0.010} \\
\midrule
 CYBORG \cite{boyd2021cyborg} & ResNet & \textbf{0.703\tiny{$\pm$0.005}} & 0.887\tiny{$\pm$0.003} & 0.745\tiny{$\pm$0.003} \\
\midrule
Unet+Gaze \cite{karargyris2021creation} \ddag & ResNet & 0.695\tiny{$\pm$0.003} & 0.876\tiny{$\pm$0.006} & 0.741\tiny{$\pm$0.004} \\
 & DenseNet & 0.692\tiny{$\pm$0.006} & 0.855\tiny{$\pm$0.004} & 0.747\tiny{$\pm$0.005} \\
 & ConvNext & 0.214\tiny{$\pm$0.057} & 0.105\tiny{$\pm$0.092} & 0.093\tiny{$\pm$0.094} \\
 & Swin-T & 0.715\tiny{$\pm$0.015} & 0.250\tiny{$\pm$0.233} & 0.584\tiny{$\pm$0.121} \\
 \midrule
MENTOR \cite{crum2023mentor} \ddag & ResNet & 0.614\tiny{$\pm$0.007} & 0.849\tiny{$\pm$0.005} & 0.636\tiny{$\pm$0.003} \\
 & DenseNet & 0.691\tiny{$\pm$0.009} & 0.871\tiny{$\pm$0.005} & 0.747\tiny{$\pm$0.003} \\
 & ConvNext & 0.323\tiny{$\pm$0.062} & 0.329\tiny{$\pm$0.064} & 0.389\tiny{$\pm$0.025} \\
 & Swin-T & 0.701\tiny{$\pm$0.008} & 0.731\tiny{$\pm$0.018} & 0.753\tiny{$\pm$0.005} \\
 \midrule
{\bf Almost Right} & ResNet & \textbf{0.703\tiny{$\pm$0.008}} & \textbf{0.891\tiny{$\pm$0.005}} & \textbf{0.744\tiny{$\pm$0.004}} \\
{\bf (proposed)} & DenseNet & \textbf{0.710\tiny{$\pm$0.005}} & 0.885\tiny{$\pm$0.030} & 0.752\tiny{$\pm$0.002} \\
 & ConvNext & \textbf{0.757\tiny{$\pm$0.005}} & \textbf{0.864\tiny{$\pm$0.017}} & \textbf{0.767\tiny{$\pm$0.002}} \\
 & Vit-b & \textbf{0.697\tiny{$\pm$0.044}} & \textbf{0.793\tiny{$\pm$0.009}} & \textbf{0.692\tiny{$\pm$0.010}} \\
 & Swin-T & \textbf{0.779\tiny{$\pm$0.010}} & \textbf{0.973\tiny{$\pm$0.004}} & \textbf{0.823\tiny{$\pm$0.004}} \\
\bottomrule
\end{tabular}
\end{table*}

%% file: tables/alpha-open-set.tex
\begin{table*}[t]
\centering
\footnotesize
\caption{The AUROC scores illustrating an effect of balancing between regular cross-entropy component and orthogonalization component ($\alpha$ in Eq. \eqref{eq:almost-ar-loss}) in the proposed {\it Almost Right} loss on open-set classification scenarios and for an example backbone (ResNet).}
\vspace{-3mm}
\label{tab:almost-ablation-alpha}
\begin{tabular}{l|ccc}
\toprule
$\alpha$ & \textbf{Iris Presentation} & \textbf{Chest X-ray} & \textbf{Synthetic Face} \\
 & \textbf{Attack Detection} & \textbf{Anomaly Detection} & \textbf{Detection} \\
\midrule
$0.10$ & 0.890\tiny{$\pm$0.003} & 0.847\tiny{$\pm$0.003} & 0.598\tiny{$\pm$0.040} \\
$0.25$ & 0.903\tiny{$\pm$0.011} & 0.844\tiny{$\pm$0.009} & 0.546\tiny{$\pm$0.047} \\
$0.40$ & 0.905\tiny{$\pm$0.003} & \textbf{0.849\tiny{$\pm$0.005}} & \textbf{0.574\tiny{$\pm$0.033}} \\
$0.50$ & \textbf{0.913\tiny{$\pm$0.012}} & \text{0.848\tiny{$\pm$0.006}} & 0.572\tiny{$\pm$0.032} \\
$0.60$ & 0.891\tiny{$\pm$0.006} & 0.843\tiny{$\pm$0.009} & 0.526\tiny{$\pm$0.010} \\
$0.75$ & 0.897\tiny{$\pm$0.028} & 0.842\tiny{$\pm$0.008} & 0.540\tiny{$\pm$0.085} \\
$0.90$ & 0.901\tiny{$\pm$0.010} & 0.836\tiny{$\pm$0.014} & 0.554\tiny{$\pm$0.019} \\
\bottomrule
\end{tabular}
\vspace{-5mm}
\end{table*}

%% file: latex/05-Conclusion.tex
\section{Conclusion}
\label{sec:Almost-Right-Conclusion}

Inspired by the efficient signal processing found within the primary visual cortex, this paper proposes a new regularization loss component that encourages a decoupling and de-correlation of the first-layer kernels by imposing soft orthogonality, enabling more efficient feature extraction for downstream layers. The experiments show that the proposed approach aids model generalization performance, especially in open-set recognition tasks (the central focus of this study). Specifically, we demonstrated generalization gains across three open-set recognition tasks and across five different architectures, surpassing existing orthogonalization methods and SOTA saliency-based regularization methods within each respective task. We offer source codes of the proposed approach and model weights to facilitate full replicability of this work.